\pdfoutput=1

\documentclass[11pt]{article}

\usepackage{acl}

\usepackage{amsmath,amsfonts,bm}

\def\eqref#1{equation~\ref{#1}}

\def\1{\bm{1}}

\DeclareMathAlphabet{\mathsfit}{\encodingdefault}{\sfdefault}{m}{sl}
\SetMathAlphabet{\mathsfit}{bold}{\encodingdefault}{\sfdefault}{bx}{n}

\usepackage{times}
\usepackage{latexsym}
\usepackage{graphicx}
\usepackage{multirow}
\usepackage{color}
\usepackage{booktabs}
\usepackage{amsmath}
\usepackage{subcaption}
\usepackage{algorithm}
\usepackage{algorithmic}
\usepackage{hyperref}
\usepackage[T1]{fontenc}

\usepackage[utf8]{inputenc}

\usepackage{microtype}
\usepackage{pifont}

\title{Graph-Structured Speculative Decoding}

\author{ \textbf{Zhuocheng Gong\textsuperscript{1}\footnotemark[1], Jiahao Liu\textsuperscript{2}, Ziyue Wang\textsuperscript{3}, Pengfei Wu\textsuperscript{1}}\\
\textbf{Jingang Wang\textsuperscript{2}, Xunliang Cai\textsuperscript{2}, Dongyan Zhao\textsuperscript{1,4}\footnotemark[2], Rui Yan\textsuperscript{5}}\footnotemark[2] \\
\textsuperscript{1}Wangxuan Institute of Computer Technology, Peking University \\
\textsuperscript{2}Meituan; \textsuperscript{3}Tianjin University; \textsuperscript{4}National Key Laboratory of General Artificial Intelligence \\
\textsuperscript{5}Gaoling School of Artificial Intelligence, Renmin University of China \\
\texttt{\{gzhch,zhaody\}@pku.edu.cn}, \texttt{pengfeiwu1999@stu.pku.edu.cn}\\ \texttt{ruiyan@ruc.edu.cn}, \texttt{wangziyue@tju.edu.cn} \\
\texttt{\{liujiahao12,wangjingang02,caixunliang\}@meituan.com}}

\begin{document}
\maketitle
\begin{abstract} 
Speculative decoding has emerged as a promising technique to accelerate the inference of Large Language Models (LLMs) by employing a small language model to draft a hypothesis sequence, which is then validated by the LLM. The effectiveness of this approach heavily relies on the balance between performance and efficiency of the draft model.
In our research, we focus on enhancing the proportion of draft tokens that are accepted to the final output by generating multiple hypotheses instead of just one. This allows the LLM more options to choose from and select the longest sequence that meets its standards.
Our analysis reveals that hypotheses produced by the draft model share many common token sequences, suggesting a potential for optimizing computation. Leveraging this observation, we introduce an innovative approach utilizing a directed acyclic graph (DAG) to manage the drafted hypotheses. This structure enables us to efficiently predict and merge recurring token sequences, vastly reducing the computational demands of the draft model. 
We term this approach Graph-structured Speculative Decoding (GSD).
We apply GSD across a range of LLMs, including a 70-billion parameter LLaMA-2 model, and observe a remarkable speedup of 1.73$\times$ to 1.96$\times$, significantly surpassing standard speculative decoding\footnote{Code available at \href{https://github.com/gzhch/gsd}{https://github.com/gzhch/gsd}}.
\end{abstract}

\renewcommand{\thefootnote}{\fnsymbol{footnote}}
\footnotetext[1]{Work done during an internship at Meituan.}
\footnotetext[2]{Corresponding authors: Dongyan Zhao (zhaody@pku.edu.cn) and Rui Yan (ruiyan@ruc.edu.cn).}

\section{Introduction}
The impressive performance of Large Language Models (LLMs) comes with an efficiency bottleneck that hinders their broader adoption~\cite{DBLP:conf/nips/VaswaniSPUJGKP17,touvron2023llama,chatgpt,DBLP:journals/corr/abs-2307-09288}.  In this context, speculative decoding (SD) emerges as a promising direction to accelerate the decoding process by reducing the number of forward passes of LLMs~\citep{DBLP:journals/corr/abs-2302-01318,DBLP:conf/icml/LeviathanKM23,DBLP:journals/corr/abs-2310-08461,DBLP:journals/corr/abs-2308-04623,DBLP:journals/corr/abs-2305-09781}. The underlying idea of SD is ``draft then verify": rather than generating one token at a time using the LLM, SD employs a smaller model to draft a hypothesis sequence of tokens covering several decoding steps and then uses the LLM to verify the hypothesis. Consequently, the decoding process includes a \textit{draft stage} and a \textit{verification stage}. In this scheme, the number of forward calls of LLMs can be significantly reduced.

\begin{figure}[t]
  \centering
  \includegraphics[width=1.\linewidth]{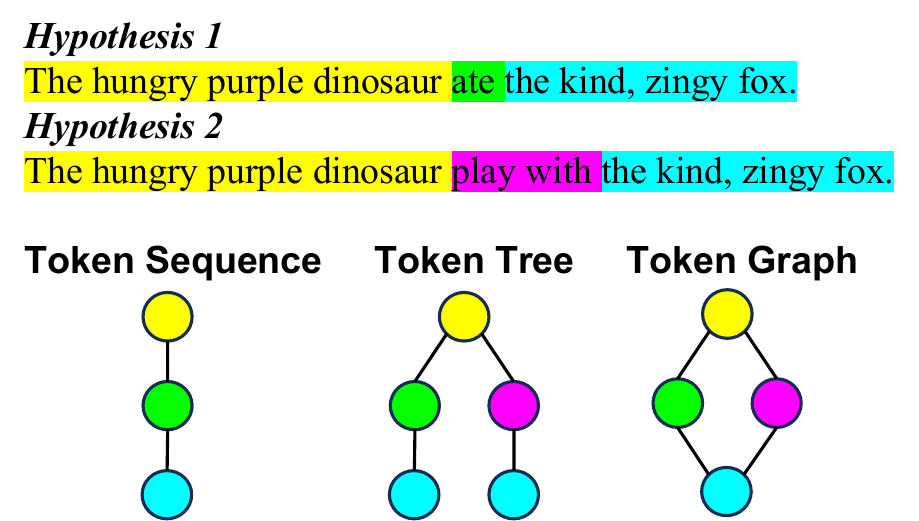}
  \caption{An illustrative comparison between the tree- and graph-structured draft token management.}
  \label{fig:intro}
\end{figure}

However, SD faces its own set of challenges: the trade-off between performance and efficiency of the draft model limits the potential for acceleration. Ideally, the draft model should generate high-quality hypotheses while maintaining computational efficiency — a balance that is notoriously difficult to strike, echoing the adage that "there's no such thing as a free lunch."
In this study, we address the challenge of enhancing the acceptance rate of the draft model's hypotheses without increasing the computational burden. Inspired by beam search~\cite{DBLP:journals/corr/abs-1211-3711} and tree attention~\cite{DBLP:journals/corr/abs-2308-04623,DBLP:journals/corr/abs-2305-09781}, our approach involves producing a bunch of hypotheses instead of a solitary one. Then, the LLM verifies these multiple hypotheses in a singlar forward pass and accepts the longest one.
While tree decoding, which adopts a tree structure to organize the drafted tokens, presents an efficient implementation for simultaneously drafting all hypotheses, it also leads to exponential growth in the number of tokens at deeper levels of the tree, resulting in a prohibitive computational overhead. Consequently, the length of the hypotheses must be kept relatively short, which in turn leads to suboptimal use of the draft model's capabilities.

Our objective is to extend the length of drafted hypotheses without a corresponding rise in computational cost. To this end, we meticulously examined the hypotheses to find opportunities for improvement. We observe that hypotheses based on the same context are often semantically similar or related, and the variations among differing hypotheses typically boil down to only a handful of tokens. Notably, more than 70\% of the drafted tokens tend to recur across various hypotheses.
If we could discern when the draft model is likely to predict these re-occurring tokens, we could simply reuse them from previous drafts, thereby reducing the overall number of tokens that need to be generated. Capitalizing on this revelation, we propose Graph-structured Speculative Decoding (GSD), which uses a directed acyclic graph to organize the drafted tokens (Figure~\ref{fig:intro}). In this graph, each path that stems from the root node corresponds to a unique hypothesis. This approach allows different hypotheses to share a substantial number of common nodes. 

The pipeline of GSD follows that of standard SD (also the Sequence-structured SD, SSD), which encompasses a draft stage and a verification stage. In the draft stage, the draft model constructs a token graph containing multiple hypotheses. In the verification stage, the token graph is flattened into a sequence, enabling the LLM to validate all hypotheses concurrently. The longest one is then adopted as part of the final output.
We conduct extensive experiments using LLaMA-70b, one of the largest open-source LLMs, showing that GSD drafts tokens not exceeding 2$\times$ the amount drafted by SSD on average, while tree-structured SD (TSD) drafted a token count that is more than 15 times greater. In terms of speedup, GSD outperforms all other methods, marking a significant advancement in speculative decoding techniques

\section{Related Works}
\subsection{LLM Compression}
Improving the efficiency of LLM inference has emerged as a pivotal research focus in recent years. The primary objective of model compression is to decrease computational demands and speed up the inference process. Research into the compression of large language models branches out into several directions, including knowledge distillation~\citep{jiao2020tinybert,sanh2019distilbert,wang2021minilmv2,passban2021alp}, quantization~\citep{tao2022compression,DBLP:journals/corr/abs-2305-17888,liu2023llm,dettmers2023qlora,xiao2023smoothquant}, network pruning~\citep{liang2021super,DBLP:conf/icml/FrantarA23}. 
Despite their innovations, these methods can be classified as lossy compression. This means that their efficiency improvements are intrinsically linked to a trade-off in performance, leading to the likelihood that a compressed LLM might produce compromised results.

\subsection{LLM Decoding Acceleration}
Alongside conventional model compression techniques, there is another branch of research that focuses on accelerating LLM inference without incurring information loss. Among these studies, speculative decoding (SD)~\citep{DBLP:journals/corr/abs-2302-01318,DBLP:conf/icml/LeviathanKM23,DBLP:journals/corr/abs-2310-08461,DBLP:journals/corr/abs-2308-04623,DBLP:journals/corr/abs-2305-09781} emerges as a promising technique. SD does not modify the model architecture, nor does it require supplemental data or retraining. SD typically employs a smaller model to draft initial predictions for ``easy" tokens, while the LLM itself verifies these drafted tokens and generates ``hard" tokens. Some researchers suggest that the smaller model is not essential for SD. For instance, the smaller model can be substituted with the LLM itself~\cite{DBLP:journals/corr/abs-2309-08168} or a large text database~\cite{he2023rest}.
In addition to SD, other efforts are being made to enhance the decoding efficiency of LLMs. Blockwise parallel decoding~\cite{DBLP:conf/nips/SternSU18}, for example, is introduced to make predictions for multiple time steps in parallel. More recently, Medusa~\cite{medusa} has trained multiple prediction heads to predict the next set of tokens simultaneously.

\begin{figure*}[t]
  \centering
  \includegraphics[width=1.\linewidth]{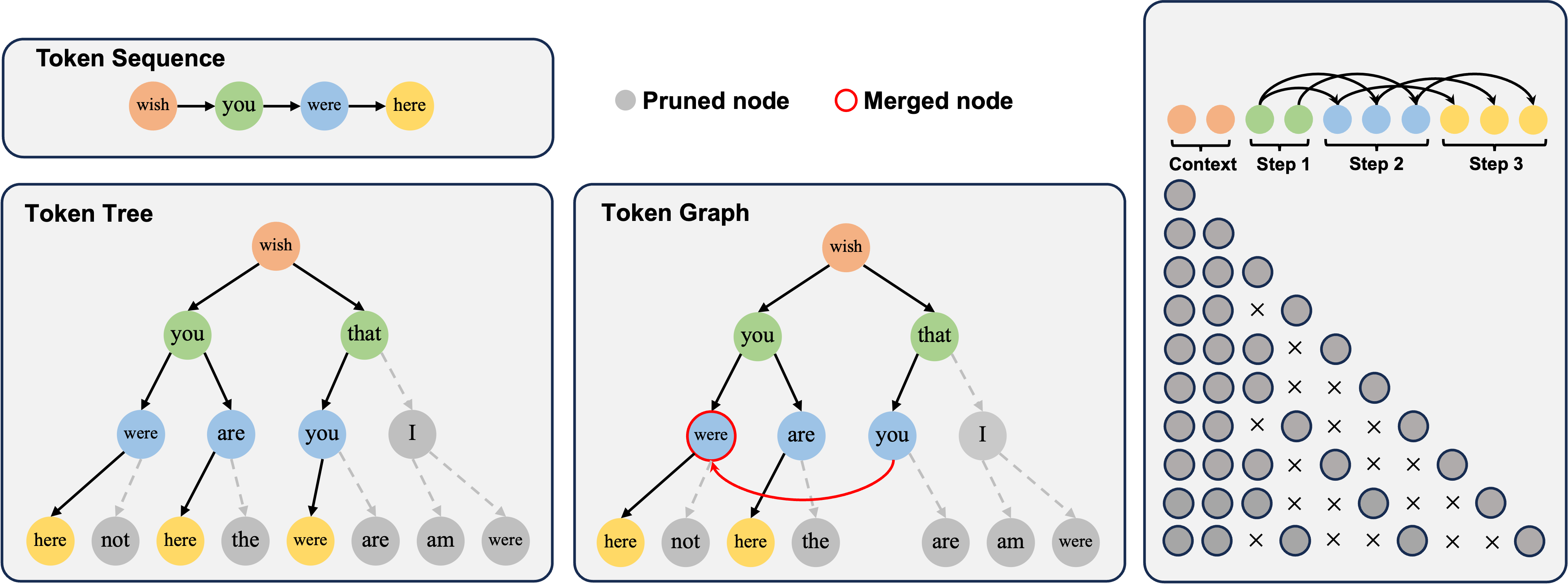}
  \caption{Overview of our method. (Left) GSD advances beyond TSD and SSD by implementing pruning strategies along with a re-occurring node merging technique. (Right) An illustration demonstrates the process by which the token tree (or graph) is flattened to a sequence. The sequence is then paired with a customized attention mask designed to uphold the proper dependencies between tokens to perform efficient drafting and verifying.}
  \label{fig:main}
\end{figure*}

\section{Preliminaries: Sequence-structured Speculative Decoding}
\label{sec:pre}
In this section, we establish the notation and provide a foundational overview of sequence-structured speculative decoding (SSD).

Consider an input sequence at time step $t$, denoted by $x_{\le t} = \{x_1, x_2, ..., x_t\}$, where each $x_i$ symbolizes the $i$-th token from the sequence. Let $M_p$ be the target LLM we want to accelerate, and let $M_q$ denote the draft model. The probabilities $p(x_{t+1}|x_{\le t})$ and $q(x_{t+1}|x_{\le t})$ represent the predictive distributions for the next token as given by $M_p$ (the LLM) and $M_q$ (the draft LM), respectively.

SSD leverages the draft model, $M_q$, to propose a hypothesis comprising $\gamma$ tokens, which we denote as $h=\{\tilde{x}_{t+1},\tilde{x}_{t+2},...,\tilde{x}_{t+\gamma}\}$. The drafting of each token, $\tilde{x}{_t+i}$, is modeled as follows:
\begin{equation}
\label{eq:1}
    \tilde{x}_{t+i} \sim q(x|x_{\le t}, \tilde{x}_{t+1},...,\tilde{x}_{t+i-1})
\end{equation} 
Upon completion of the draft stage, the LLM verifies the $\gamma$ drafted tokens in a singular forward pass. 
The verification process, which compares predictions made by $M_p$ and $M_q$ to determine which tokens shall be accepted, can be conducted in both deterministic and non-deterministic ways.
Deterministic verification accepts drafted tokens only if the LLM would generate the same.
The non-deterministic way employs the sampling method used in previous studies~\cite{DBLP:journals/corr/abs-2302-01318}. For the $i$-th token in the hypothesis, the acceptance probability is calculated as $\min(1,p(\tilde{x}_{t+i})/q(\tilde{x}_{t+i}))$. 
Should the token $\tilde{x}_{t+i}$ face rejection, all subsequent tokens in the hypothesis are also discarded, the verification process comes to a halt, and $M_p$ regenerates the discarded token. 
This method ensures that the tokens that are ultimately accepted are representative of the output distribution characterized by $M_p$.


\section{A Step Forward: Tree-structured Speculative Decoding}
\label{sec:treesd}
An intuitive idea for improving SSD is to draft multiple hypotheses instead of merely one. This is where Tree-structured SD (TSD) comes into play.

In each drafting step of SSD, the draft model predicts a single next token as described in Equation~\ref{eq:1}. After $\gamma$ steps, the drafted tokens compose a sequence $\{\tilde{x}_{t+1},\tilde{x}_{t+2},...,\tilde{x}_{t+\gamma}\}$. In contrast, TSD allows the draft model to consider $k$ different alternatives for the next token at each drafting step. The resulting drafted tokens thus create a tree structure, with the root representing the context at the commencement of drafting, and each branch from the root depicting a different hypothesis.

After $\gamma$ drafting steps, the resulting token tree has a depth of $\gamma$ and a maximum out-degree of $k$ and can contain up to $\frac{k^{\gamma+1}-1}{k-1}$ nodes, representing as many as $k^\gamma$ unique hypotheses.
Let's denote the collection of all hypotheses as $\{h_i\}_{i=1}^{k^\gamma}$.
TSD holds a significant advantage over SSD; by enabling the generation of a larger pool of hypotheses in a single drafting stage, it raises the chances of having longer sequences of tokens accepted by the LLM. This boosts the acceptance rate of the SD process. Fundamentally, TSD operates in a manner analogous to beam search, maintaining multiple potential hypotheses within its tree structure during the draft stage and then selecting the most promising one during the verification stage.

\subsection{Parallelized drafting and verifying via tree attention}
The draft stage of TSD generates a multitude of hypotheses. A significant challenge within this framework is the efficient drafting of these multiple hypotheses. If one were to adhere to the traditional inference scheme that decodes one token at a time (akin to extending one branch of the token tree), the computational demands are apparently unacceptable given that the token tree contains $\frac{k^{\gamma+1}-1}{k-1}$ tokens to be decoded.

A promising resolution to this problem is by employing meticulous tree attention. Tree attention operates by flattening the token tree into a sequence and then simultaneously predicting the next node for all branches during a single forward draft, thus circumventing the necessity of performing a forward pass for each potential sequence. As illustrated in Figure~\ref{fig:main}, it accomplishes this by customizing the attention mask in such a way that each token is only allowed to attend to its ancestor nodes in the tree hierarchy, thus maintaining the correct dependencies amongst tokens.

The verification stage benefits from tree attention by validating all hypotheses within a single forward pass. After this process, the longest path that unfolds from the root node is chosen as the sequence to be accepted. 

\subsection{Pruning inferior branches}
Despite the parallel drafting and verification with tree attention, TSD still consumes significantly more computation than SSD. The root cause lies in the exponentially increased length of input sequences processed in each forward pass. Transformer attention has a computational complexity that scales quadratically, $\mathcal{O}(l^2)$, with the sequence length $l$. While kv-caching does alleviate the computational load to some degree, the burden remains substantially heavier than that of SSD. 
Thus, to reduce the input sequence length, we need to perform pruning on the token tree.

We introduce two pruning strategies to moderate the size of the token tree. The first strategy is \emph{probability pruning}. For a given node $c$ within the token tree, where $s_c$ denotes the path from the root to $c$, the logit probability is given by $q(c|x_{\le t}, s_c)$. By setting a probability threshold $\theta_{prob}$, we can filter out nodes: if $q(c|x_{\le t}, s_c) < \theta_{prob}$, the node is deemed unlikely to be verified successfully and is marked as a leaf, halting further speculation.

The second strategy, \emph{sibling pruning}, focuses on a node's child nodes $\{c_i\}_{i=1}^k$. Among these, we discern which nodes should remain as non-leaf nodes based on their logit probabilities relative to the highest probability among them. Specifically, let $m_q = \max_{i=1,...,k}p(c_i|x_{\le t}, s_{c_i})$. A child node $c_i$ is then designated as a leaf if $p(c_i|x_{\le t}, s_{c_i}) < \theta_{sib} \cdot m_q$.
This approach ensures that the logit probabilities among sibling nodes do not deviate excessively from the maximum observed, $m_q$. The underlying idea is that, during probabilistic sampling, if the generation probabilities across a node's children vary greatly, the tokens associated with lower probabilities are less likely to be chosen. Therefore, it may not be necessary to keep these less probable nodes in the tree. Hence, when the output distribution for a current token is peaked—indicating high model confidence in its prediction—we need not preserve many child nodes. However, if the distribution is flatter, meaning multiple tokens have similar probabilities, it then becomes prudent to maintain a broader set of child nodes as candidates.

\section{Graph-structured Speculative Decoding}
Empirically, we observe that TSD often fails to surpass SSD, contrary to expectations. It appears that despite the utilization of pruning and tree attention, the cost of drafting multiple hypotheses still counterbalances the potential benefits that TSD offers. So we would like to ask: Can we further reduce the quantity of drafted tokens to enhance TSD's efficiency and effectiveness?

\begin{figure}[t]
  \centering
  \includegraphics[width=1.\linewidth]{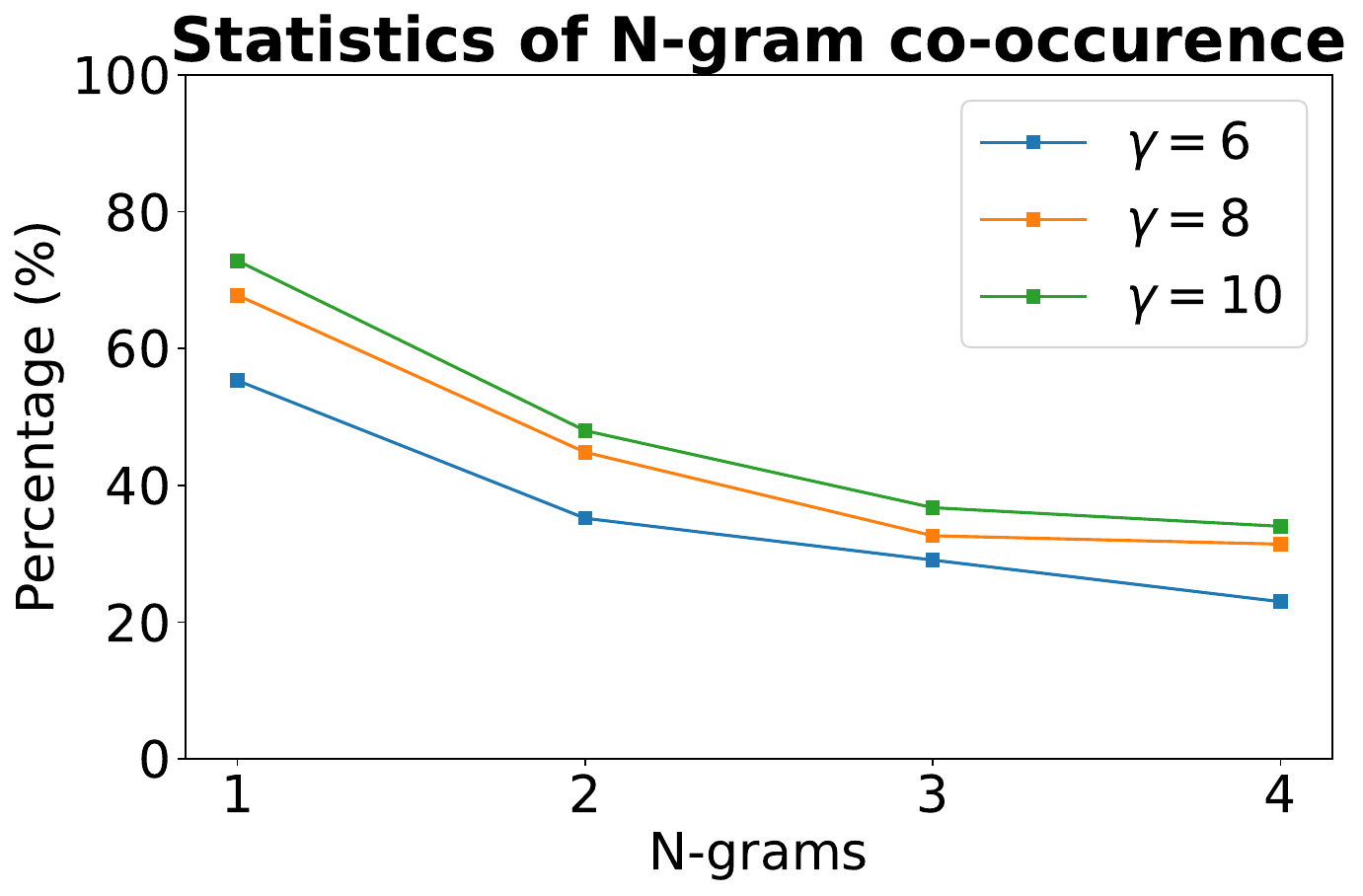}
  \caption{The proportion of tokens that are part of re-occurring n-grams within the token tree where the maximum out-degree $k$ is 4. $\theta_{prob} = 0.2$ and $\theta_{sib}=0.3$.}
  \label{fig:stat}
\end{figure}

\subsection{Same tokens re-occur among hypotheses}
Before delving into GSD, we first conduct a pilot study to investigate the drafted hypotheses generated by TSD. We analyze the token trees from 100 distinct TSD runs, documenting the statistics of n-gram co-occurrences across various branches. The findings of this analysis are presented in Figure~\ref{fig:stat}, and they give rise to several key insights:
\begin{itemize}
    \item There is a high degree of commonality among the tokens in different hypotheses. As depicted in Figure~\ref{fig:stat}, within a token tree of 10-depth and 4-width, approximately 70\% of tokens appear across multiple branches. This suggests that the generated hypotheses tend to form a cluster of semantically similar or related candidates, rather than branching off in completely disparate semantic directions. 
    \item There is also a notable frequency of recurring n-grams within the token tree. This observation suggests that the similarities between different hypotheses extend beyond single tokens — entire segments of tokens (n-grams) are often duplicated among the various branches of the tree. This pattern points to redundancy in the token sequences being drafted, which may have implications for optimizing the efficiency of the speculative decoding process.
\end{itemize}


\subsection{Identifying redundant nodes}
\label{sec:identify}
We leverage the findings of identical tokens reappearing across different hypotheses to reduce computation.
To this end, we introduce the concept of a $\tau$-redundant node. A node is designated as $\tau$-redundant when it corresponds to the last token of a re-occurring $\tau$-gram. We assume that the presence of a $\tau$-gram, defined as a sequence of $\tau$ consecutive identical tokens, signals a high degree of similarity between the current hypothesis and an alternate hypothesis already explored. This implies a strong likelihood that the sequence will continue to predict identical subsequent tokens.



\subsection{Merging redundant nodes}
\label{sec:merge}
Building on the concept of $\tau$-redundant nodes, we implement a procedure to merge these nodes to enhance efficiency. The approach is straightforward: we mark $\tau$-redundant nodes as leaf nodes, effectively ceasing their further expansion within the token tree.
To merge the nodes, we first locate the first occurrence of the re-occurring $\tau$-gram. We then draw a directed edge from the $\tau$-redundant node to this first occurrence. By doing so, we establish that the nodes following the $\tau$-redundant node will not need to be generated anew. Rather, we can directly reuse the results previously computed for the initial $\tau$-gram occurrence. As a result of this merging process, the token tree is transformed into a directed acyclic graph (DAG), wherein no n-grams longer than $\tau$ will be repeated.

\paragraph{How does node merging hurt the performance?}
Merging nodes can result in a divergence from the nodes that would have otherwise been generated, potentially impacting the quality of the generated content. To quantify this effect, we calculate the KL divergence between the probability distributions of the next token across the vocabulary with or without node merging. Experimental results demonstrate that the KL divergence decreases rapidly with the increase of $\tau$, suggesting that the impact of node merging diminishes significantly as the threshold $\tau$ is heightened. (Detailed results in Appendix~\ref{ap:tau})


\begin{figure}[t]
  \centering
  \includegraphics[width=1.\linewidth]{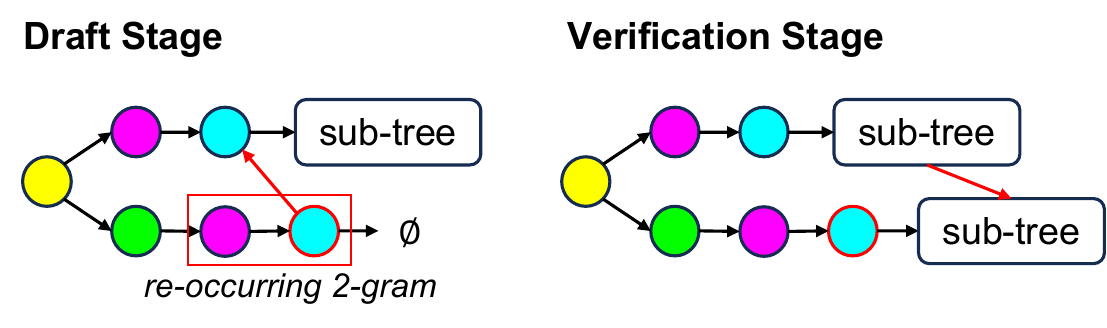}
  \caption{An illustration of how the token graph operates during the draft stage and the verification stage.}
  \label{fig:verify}
\end{figure}

\subsection{Token graph verification}
There is still one step to go to fulfill GSD: the verification process. In the verification stage, we need to flatten the token graph to a sequence so that the LLM can verify all hypotheses simultaneously. To convert a DAG into a sequence while preserving the correct dependencies between tokens, we start by reverting the graph to its original tree structure. This is done by ``unmerging" all previously merged nodes. During this process, the successor nodes of any redundant node are replicated from the relevant merged nodes (Figure~\ref{fig:verify}). With the structure now back in the form of a tree, we can apply the same verification procedure as used in TSD.

\begin{table*}[t]
\begin{center}
\begin{tabular}{lllllllllr}
\toprule
\textbf{Datasets} & \textbf{Mehtod} & \textbf{Model} & \textbf{Acceptance} &  \textbf{Drafted} & \textbf{Graph} & \textbf{Speedup} \\
&&  & \textbf{Rate} & \textbf{Token Num} & \textbf{Success}  \\
\midrule
GSM8k & Self SSD & LLaMA-2-70b & - & - & - & 1.37$\times$\\
GSM8k & SSD & LLaMA-2-70b & 0.795 & 629.9 & - &1.85$\times$  \\
GSM8k & TSD & LLaMA-2-70b & 0.894 & 8574.6 & 0\% & 1.81$\times$  \\
GSM8k & GSD & LLaMA-2-70b & 0.917 & 793.1 & 27.7\% &\textbf{1.96}$\times$ \\ \midrule
XSUM & Self SSD & LLaMA-2-70b & - & - & - & 1.28$\times$ \\
XSUM & SSD & LLaMA-2-70b & 0.652 & 773.2 & - &1.56$\times$  \\
XSUM & TSD & LLaMA-2-70b & 0.784 & 22512.4 & 0\% & 1.42$\times$  \\
XSUM & GSD & LLaMA-2-70b & 0.831 & 1544.8 & 32.8\% &\textbf{1.73}$\times$  \\
XSUM & SSD & LLaMA-2-70b-chat & 0.496 & 989.4 & - & 1.19$\times$  \\
XSUM & TSD & LLaMA-2-70b-chat & 0.634 & 4601.2 & 0\% & 1.30$\times$  \\
XSUM & GSD & LLaMA-2-70b-chat & 0.642 & 1545.7 & 30.4\% & \textbf{1.32}$\times$  \\
\bottomrule
\end{tabular}
\end{center}
\caption{Evaluation results on 70b model. Self SSD is the method proposed by \citet{DBLP:journals/corr/abs-2309-08168}, which uses the LLM itself as the draft model. Speedup is the averaged result of greedy and top-$p$ sampling. Here we only present the results of 70b models, full results can be found in Appendix~\ref{ap:deter}.}
\label{tab:main}
\end{table*}

\section{Experiments}
\subsection{Setup}
There are two settings for verifying the drafted tokens: a deterministic setting where accepting the drafted tokens only if the LLM would generate tokens the same, and a non-deterministic setting where accepting the drafted tokens if they follow the same distribution with the LLM-itself generated tokens.
In our main experiments, we adhere to the deterministic decoding setting if not specified. Under this condition, the generated output sequence is guaranteed to be identical to what would be produced via standard generation methods, so we can concentrate solely on efficiency metrics. Other details can be found in Appendix~\ref{ap:cfg}.


\paragraph{Models}
We experiment on various backbone LLMs, including LLaMA~\cite{touvron2023llama}, OPT~\cite{zhang2022opt}, and BLOOM~\cite{workshop2022bloom}. For LLaMA, we use LLaMA-70b, LLaMA-70b-chat, and LLaMA-7b as large LLMs and LLaMA-7b and LLaMA-7b-chat, LLaMA-160m as draft models respectively. Note that LLaMA-160m is not an official checkpoint but a LLaMA-like model~\cite{miao2023specinfer}. For OPT, we use OPT-13b as the LLM and OPT-350m as the draft model. For BLOOM, we use BLOOM-7b1 as the LLM and BLOOM-560m as the draft model.

\paragraph{Datasets}
We evaluate on Extreme Summarization (XSum)~\cite{DBLP:conf/emnlp/NarayanCL18}, GSM8K~\cite{DBLP:journals/corr/abs-2110-14168}, Alpaca~\cite{alpaca}, and WMT-14 (En-De)~\cite{bojar-EtAl:2014:W14-33}. For GSM8K and WMT-14, we evaluate the full test set. For XSum and Alpaca, we randomly select 5000 instances for evaluation. 

\subsection{Main Results}
Table~\ref{tab:main} illustrates a comparison of our method against other speculative decoding approaches. Focusing on the speed-up ratio, we can see that GSD offers a significant advantage over the alternatives, achieving up to 1.94 and 1.70 times faster speeds. 
When examining the acceptance rate, we observe that both TSD and GSD have an acceptance rate that exceeds that of SSD by more than 10\%. This indicates that tokens generated by the draft model are more likely to pass the verification process. 
Comparing the number of drafted tokens, we can see that TSD produces an order of magnitude more tokens than SSD. Hence, while TSD also has a high acceptance rate, this advantage is negated by the excessive number of tokens generated.

Additionally, we assess what proportion of tokens, which passed verification during the speculative decoding process, contained nodes from the merged subtrees, and find that approximately 30\% of the drafting stages include such tokens. This indicates that, while the token graph is significantly smaller in node count compared to the token tree, we have successfully preserved the decoding information by recognizing and grafting nodes from different branches.


\begin{figure*}[t]
  \centering
  \includegraphics[width=1.\linewidth]{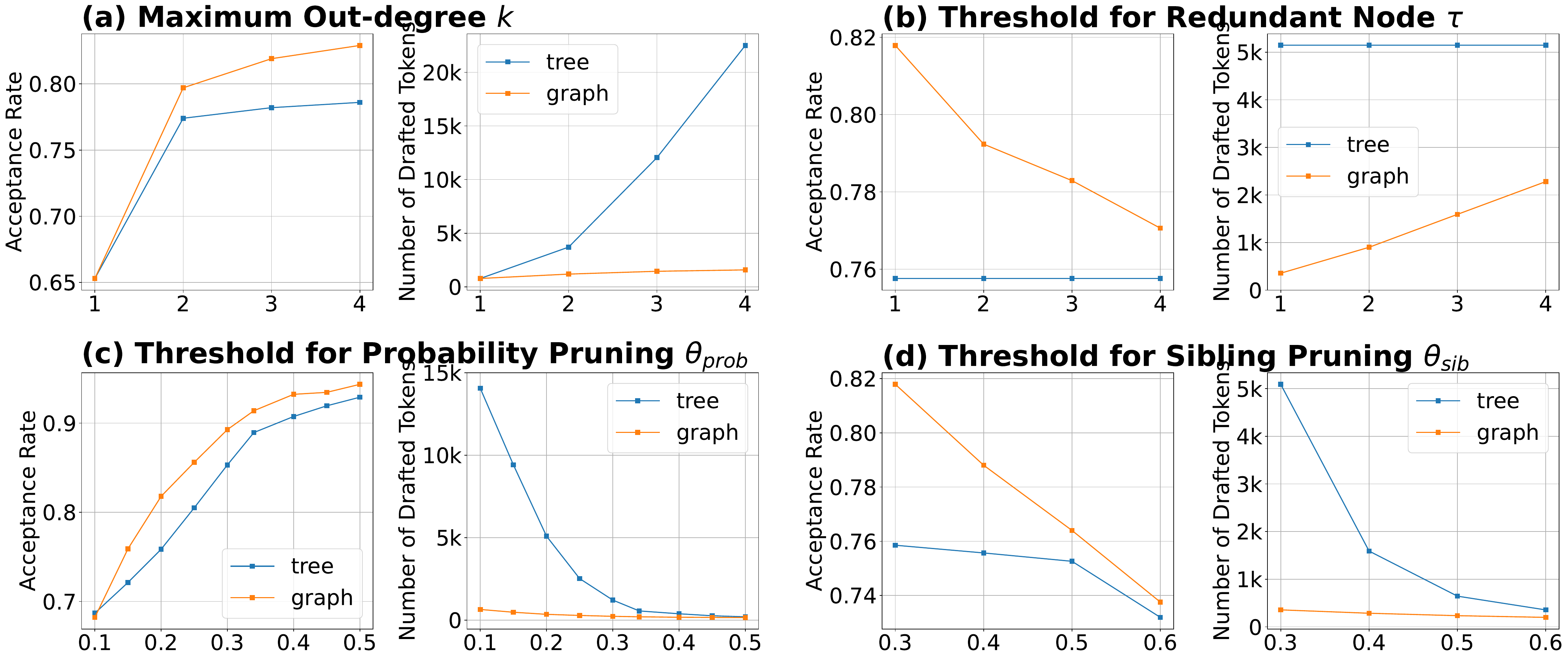}
  \caption{A series of ablation studies to investigate the hyperparameter configuration of maximum out-degree, redundant threshold, and two pruning techniques. All other hyperparameters adhere to the configuration described in section~\ref{ap:cfg}.}
  \label{fig:ab}
\end{figure*}

\subsection{Ablation Study}
\label{sec:ab}
\paragraph{Maximum Out-degree $k$}
Maximum out-degree $k$ refers to the maximum number of child nodes that each node within the token tree (or graph) can possess. As depicted in Figure~\ref{fig:ab}(a), as the $k$ increases, the model is more likely to accept longer sequences in the verification stage due to the more diverse set of candidate hypotheses, thereby significantly enhancing the acceptance rate. 
However, the total number of nodes in the token tree increases exponentially as the increase of $k$ as we have discussed in Section~\ref{sec:treesd}. When setting $k$ to 4, the token tree contains more than 20000 tokens which leads to a heavy computation budget.
In contrast, the token graph prevents the uncontrolled swell of node count that could impede computational efficiency by merging repeating sub-trees. This optimization allows the GSD to achieve a much higher acceptance rate while free from a rapid increase in nodes with the increase of $k$.

\begin{table}[t]
\begin{center}
\begin{tabular}{lllr}
\toprule
\textbf{Methods} & SSD & TSD & GSD\\
\midrule
GSM8k & 1.80$\times$ & 1.81$\times$ & 2.14$\times$\\
XSUM & 1.58$\times$ & 1.46$\times$ & 1.89$\times$\\
\bottomrule
\end{tabular}
\end{center}
\caption{Speedup results on non-deterministic speculative decoding on LLaMA-2-70b.}
\label{tab:non}
\end{table}

\paragraph{Threshold for Redundant Node $\tau$}
As mentioned in Section~\ref{sec:identify}, when two different hypotheses emanating from different branches share a common token sequence of length $\tau$, they are identified as repetition and subsequently merged as a single branch. Thus, the larger the $\tau$, the more radical the node merging becomes. As shown in Figure~\ref{fig:ab}(b), as the increase of $\tau$, the method becomes more conservative in fusing repeated branches, retaining more nodes in the token graph.
Besides, the acceptance rate is inversely correlated with the redundant threshold. This implies that more aggressive node fusion leads to a more diverse set of candidate hypotheses.
At first glance, this might seem paradoxical, since one would expect that aggressive node fusion, which reduces the number of nodes in the token graph, would decrease the diversity of hypotheses by merging similar sequences. However, when the merging happens, the two nodes that are merged as one then share a common child subtree in later drafting steps.
By merging, the newly generated tokens within the subtree are simultaneously added to two different branches, while these tokens might not be generated by both independent branches if not merged.
Thus, the node merging effectively introduces a greater variety of hypotheses by allowing for increased sharing of information between different parts of the token graph, which might otherwise remain isolated, leading to less efficient search space coverage.

\paragraph{Pruning Threshold $\theta_{prob}, \theta_{sib}$}
The probability pruning technique prunes tokens of low logit probability and the sibling pruning technique involves pruning sibling nodes that had passed the probability-based pruning based on the maximum logit probability. As illustrated in the figure, both pruning strategies significantly reduce the number of generated tokens.
However, these two pruning strategies have opposite effects on the acceptance rate. When the threshold is raised, probability pruning leads to an increase in the acceptance rate, while sibling pruning has a diminishing effect. This indicates that while probability pruning can help in focusing the speculative decoding process on more likely hypotheses, sibling pruning might lead to the removal of potential candidate hypotheses that could have been valid.
The implications of these findings suggest that a delicate balance must be struck between pruning enough to maintain computational efficiency and avoiding overly aggressive pruning that could eliminate valid hypotheses. 

\subsection{Non-deterministic Setting}
Table~\ref{tab:non} represents performance under the non-deterministic decoding setting. This non-deterministic verification process determines whether a drafted token should be accepted by comparing the generating probability of the draft model and the LLM.
Implementing GSD in this setting is a little tricky because GSD uses a shared logit distribution for redundant tokens, which could slightly deviate from the actual distribution. We have addressed the potential effects of this issue in Section~\ref{sec:merge} through experimental analysis. 
Furthermore, we conduct an explicit evaluation of the text quality, confirming that the performance disruption due to node merging is inconsequential. Detailed results can be found in Appendix~\ref{ap:non}.


\section{Analysis}
\subsection{Breakdown of Computation}
Table~\ref{tab:time} presents a computational analysis comparing different speculative decoding methods. 
Compared to TSD, the primary improvement offered by GSD lies in the reduction of time consumed during the draft stage, which can be attributed to the fewer number of nodes in the token graph, resulting in a reduced count of tokens that need to be processed during each drafting forward pass.

Besides, we find that, in addition to drafting and verifying, there is a significant portion of computation that should not be overlooked. We find that this computation is primarily associated with the update of the kv-cache of the draft model. Thus, improving the efficiency of the kv-caching represents a potential direction for further accelerating the speculative decoding. 

\begin{table}[t]
\begin{center}
\begin{tabular}{llllr}
\toprule
\textbf{Methods} & \textbf{Draft} & \textbf{Verification} & \textbf{Others}\\
\midrule
SSD & 224.9 ms & 133.5 ms & 45.8 ms \\
TSD(k=2) & 257.0 ms & 172.4 ms  & 46.9 ms \\
GSD(k=2) & 225.9 ms & 170.0 ms & 45.5 ms \\
TSD(k=4) & 323.9 ms & 184.4 ms & 49.8 ms \\
GSD(k=4) & 209.0 ms & 178.3 ms & 50.2 ms \\

\bottomrule
\end{tabular}
\end{center}
\caption{Breakdown of computation of a \textbf{single} draft-verification iteration.}
\label{tab:time}
\end{table}

\subsection{Case Study}
\begin{figure}[t]
  \centering
  \includegraphics[width=1.\linewidth]{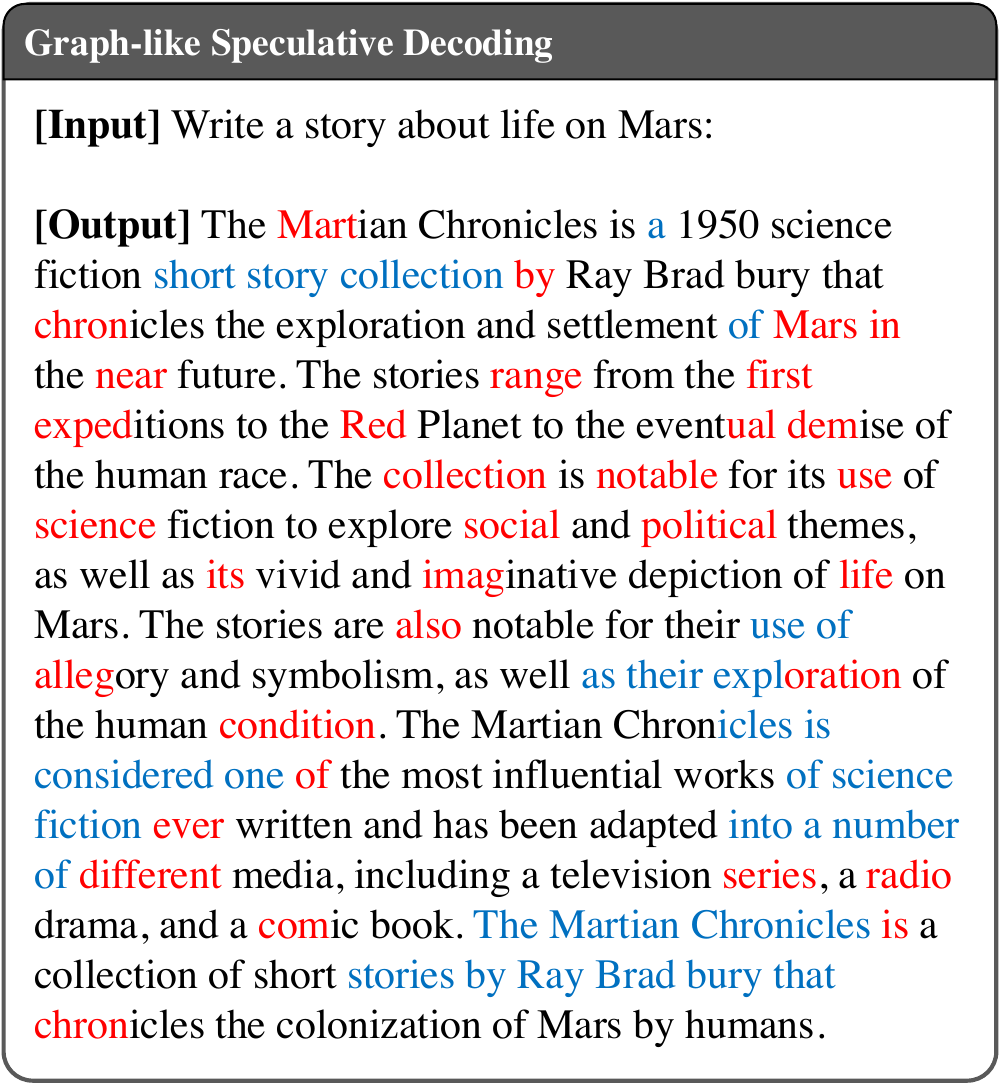}
  \caption{A visualization of the generation process of graph-structured speculative decoding. The black color represents the token generated by the verification model. Both red and blue are the accepted tokens. Red tokens are ordinarily drafted while blue tokens are from the merged nodes of the token graph.}
  \label{fig:case}
\end{figure}

Figure~\ref{fig:case} presents an illustrative example of GSD.
This case demonstrates how the token graph assists in maintaining various hypotheses while simultaneously decreasing the total number of drafted tokens. Notably, approximately 30\% of the accepted drafted tokens are derived from the subtrees associated with merged nodes, illustrating the efficiency gains achieved through GSD.

\section{Conclusion}
In this paper, we introduce graph-structured speculative decoding (GSD), a novel decoding strategy that utilizes a token graph to concurrently record a multitude of sequence hypotheses within a single draft stage. We propose a redundant node merging technique and two pruning strategies to constrain the size of the token graph without unduly compromising the diversity of hypotheses. Our extensive experiments demonstrate that GSD significantly increases the acceptance rate of drafted tokens while not introducing much computation, achieving a noticeable acceleration in speed compared to previous speculative decoding methods.

\section*{Limitations}
We discuss the limitations of our work as follows:
While our investigation has highlighted an interesting phenomenon of hypotheses generated from the same context contexts, we have not thoroughly examined the underlying mechanism that gives rise to this phenomenon. A deeper exploration into why these hypotheses exhibit such close semantic ties could unveil further insights that may benefit future research and applications.

\section*{Acknowledgement}

\bibliography{anthology,custom}
\bibliographystyle{acl_natbib}
\clearpage

\appendix
\section{Additional Implementation Details}
\label{ap:cfg}
We establish both the maximum input sequence length and output sequence length at 512. Any input sequences exceeding 512 tokens are truncated. We set the maximum drafting step at 10 and adopt a draft-exiting mechanism to prematurely exit the drafting stage when the token probability drops below $\theta_{prob}$. For the top-$p$ sampling decoding, we set the top-$p$ to 0.7 and temperature to 0.7. For graph decoding and tree decoding, we set the maximum out-degree $k$ as 4. For the pruning configurations, we default to $\theta_{prob} = 0.2$ and $\theta_{sib} = 0.3$ . We set $\tau = 2$. 

\section{Comparison with Other Inference Acceleration Methods}
\begin{table}[h]
    \centering
    \begin{tabular}{ccc}
    \toprule
       Methods  & GSM8K & XSUM \\
       \midrule
        Medusa & 2.01$\times$ & 1.62$\times$\\
        GSD & 1.96$\times$ & 1.73$\times$ \\
    \bottomrule
    \end{tabular}
    \caption{Speedup ratios on LLaMA-2-70b.}
    \label{tab:ap_medusa}
\end{table}
Except for speculative decoding, there have been other methods for accelerating the decoding of LLM. Among these studies, Medusa~\cite{medusa} is a simple yet effective method. We compare with Medusa on GSM8K and XSUM (Table~\ref{tab:ap_medusa}). Besides, we want to mention that Medusa is dedicated to the same deterministic setting and employs a similar tree structure to manage the generated tokens, so it is possible to incorporate Medusa with our proposed graph structure to further optimize the token management. Hopefully, this would bring further acceleration.

\section{Impact of Node Merging on Logits Distribution}
\label{ap:tau}
\begin{table}[h]
    \centering
    \resizebox{\linewidth}{!}{\begin{tabular}{lccccc}
    \toprule
    maximum out-degree & $\tau=0$ & $\tau=1$ & $\tau=2$ & $\tau=3$ & $\tau=4$\\
     \midrule
    k=3 & 1.19e-4 & 2.70e-6 & 5.77e-7 & 3.34e-7 & 5.13e-7 \\
    k=5 & 1.78e-4 & 4.70e-6 & 4-21e-7 & 7.62e-7 & 8.49e-7 \\
    k=$\infty$ & 1.30e-4 & 3.11e-6 & 1.03e-6 & 9.27e-7 & 7.64e-7 \\
    \bottomrule
    \end{tabular}}
    \caption{Averaged KL-divergence between the probability distributions across the vocabulary with or without node merging. Results are averaged over 1000 examples. We test on a series of $k$ (maximum out-degree) and $\tau$ (the threshold for redundant node), showing that in most cases, merging redundant nodes brings minimal affection to the generation probability of subsequent tokens.}
\end{table}

\section{Additional Results on Deterministic Setting}
We present the evaluation results on BLOOM-7b1, OPT-13b, and LLaMA-7b in Table~\ref{tab:ap_det_1},\ref{tab:ap_det_2}, and \ref{tab:ap_det_3}.
\label{ap:deter}
\begin{table}[h]
    \centering
    \resizebox{\linewidth}{!}{\begin{tabular}{lccc}
    \toprule
        Methods & Alpaca & WMT-14 en-de & gsm8k \\
        \midrule
        SSD & 0.628 (1.12$\times$) & 0.705 (1.30$\times$) & 0.653 (1.18$\times$) \\
        TSD & 0.783 (0.44$\times$) & 0.798 (0.59$\times$) & 0.741 (0.32$\times$)\\
        GSD & 0.819 (1.48$\times$) & 0.812 (1.52$\times$) & 0.755 (1.26$\times$)\\
        \bottomrule
    \end{tabular}}
    \caption{BLOOM-7b1 performance under $k=4$, $\tau=1$, $\theta_{prob}=0.4$, $\theta_{sib}=0.1$. BLOOM-560m serves as the draft model.}
    \label{tab:ap_det_1}
\end{table}

\begin{table}[h]
    \centering
    \resizebox{\linewidth}{!}{\begin{tabular}{lccc}
    \toprule
        Methods & Alpaca & WMT-14 en-de & gsm8k \\
        \midrule
        SSD & 0.563 (1.12$\times$) & 0.621 (1.16$\times$) & 0.602 (1.08$\times$) \\
        TSD & 0.672 (0.37$\times$) & 0.705 (0.38$\times$) & 0.770 (0.62$\times$)\\
        GSD & 0.691 (1.15$\times$) & 0.733 (1.28$\times$) & 0.793 (1.22$\times$)\\
        \bottomrule
    \end{tabular}}
    \caption{OPT-13b performance under $k=4$, $\tau=1$, $\theta_{prob}=0.4$, $\theta_{sib}=0.1$. OPT-350m serves as the draft model.}
    \label{tab:ap_det_2}
\end{table}

\begin{table}[h]
    \centering
    \resizebox{\linewidth}{!}{\begin{tabular}{lccc}
    \toprule
        Methods & Alpaca & WMT-14 en-de & gsm8k \\
        \midrule
        SSD & 0.729 (1.22$\times$) & 0.783 (1.29$\times$) & 0.601 (1.04$\times$)\\
        TSD & 0.846 (0.65$\times$) & 0.851 (0.56$\times$) & 0.775 (0.60$\times$)\\
        GSD & 0.860 (1.31$\times$) & 0.863 (1.36$\times$) & 0.793 (1.16$\times$)\\
        \bottomrule
    \end{tabular}}
    \caption{LLaMA-2-7b performance under $k=4$, $\tau=1$, $\theta_{prob}=0.4$, $\theta_{sib}=0.1$. LLaMA-160m serves as the draft model.}
    \label{tab:ap_det_3}
\end{table}

\section{Additional Results on Non-deterministic Setting}
\label{ap:non}
Table~\ref{tab:ap_non} shows results on LLaMA-2-7b under the non-deterministic setting. In this scenario, the text produced by the model is not necessarily identical to that which would be generated via a standard decoding process. Consequently, to ensure that GSD does not significantly impair output quality, we assess the quality of the generated text. The results are shown in Table~\ref{tab:ap_non_2}.

\begin{table}[h]
    \centering
    \resizebox{\linewidth}{!}{\begin{tabular}{lccc}
    \toprule
        Methods & Alpaca & WMT-14 en-de & gsm8k \\
        \midrule
        SSD & 0.695 (1.16$\times$) & 0.737 (1.21$\times$) & 0.540 (0.96$\times$) \\
        GSD & 0.793 (1.34$\times$) & 0.848 (1.31$\times$) & 0.836 (1.18$\times$)\\
        \bottomrule
    \end{tabular}}
    \caption{LLaMA-2-7b performance under $k=4$, $\tau=1$, $\theta_{prob}=0.4$, $\theta_{sib}=0.1$. The hyperparameter settings might not be optimal.}
    \label{tab:ap_non}
\end{table}

\begin{table}[h]
    \centering
    \resizebox{\linewidth}{!}{\begin{tabular}{lccc}
    \toprule
          & Rouge-1 & Rouge-2 & Rouge-l \\
        \midrule
        vanilla decoding  & 0.25&0.09&0.19\\
        SSD & 0.24&0.09&0.19\\
        GSD ($\tau=1$) & 0.23&0.09&0.18\\
        GSD ($\tau=2$) & 0.23&0.09&0.18\\
        \bottomrule
    \end{tabular}}
    \caption{Rouge-1/2/l scores on LLaMA-2-7b under non-deterministic setting.}
    \label{tab:ap_non_2}
\end{table}

\section{Further Analysis on GSD}
We present some extra explorations in this section.
GSD introduces a novel directed acyclic graph structure to manage the drafted tokens. Every branch starting from the root node forms a unique hypothesis. We analyze the positional structure of the accepted/rejected nodes within the graph. 

\begin{figure}[h]
  \centering
  \includegraphics[width=1.\linewidth]{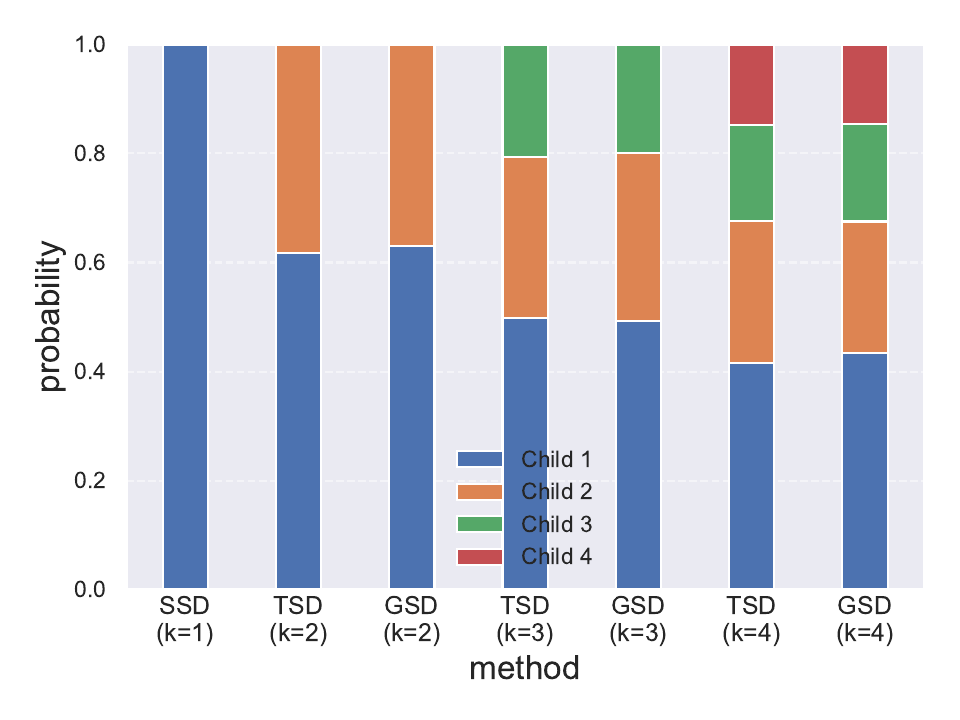}
  \caption{Percentage of $i$-th child being accepted. Results are averaged across all nodes within the token graph. We compare various speculative decoding configurations on LLaMA-7b. The child nodes within the decoding graph are ranked according to their probability, such that Child-1 corresponds to the token with the highest probability, while Child-k represents the token with the $k$-th highest logit probability.}
  \label{fig:ap1}
  \vspace{-3mm}
\end{figure}

\begin{figure}[h]
  \centering
  \includegraphics[width=1.\linewidth]{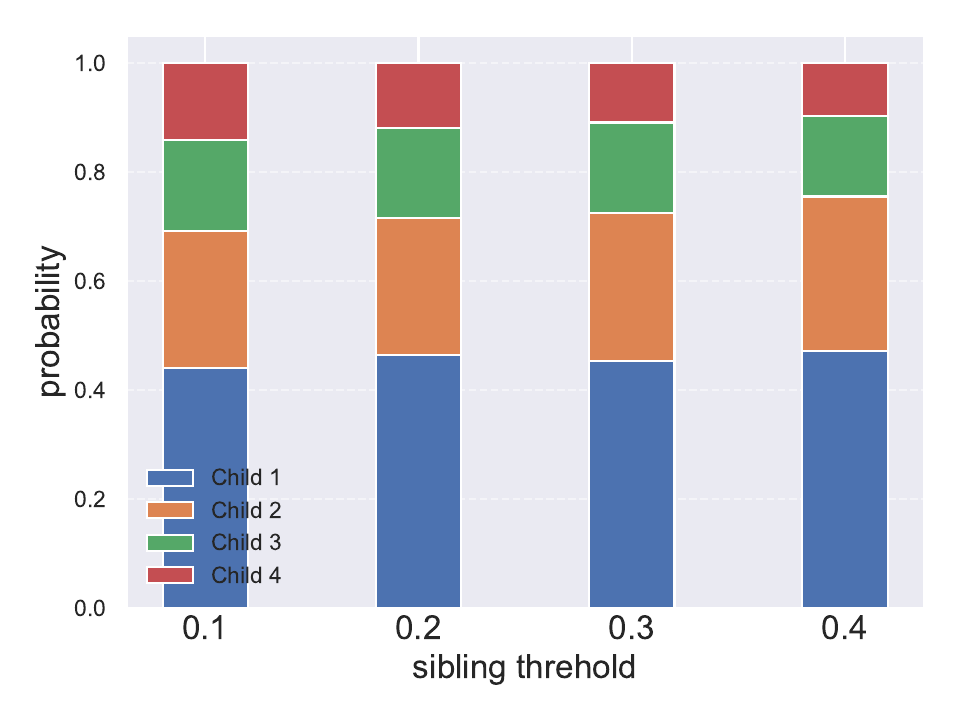}
  \caption{Percentage of $i$-th child being accepted. Results are obtained from LLaMA-7b with $k=4$, $\tau=1$, $\theta_{orob}=0.4$}.
  \label{fig:ap2}
  \vspace{-3mm}
\end{figure}

Figure~\ref{fig:ap1} shows the benefit of considering multiple hypotheses in enhancing the acceptance rate, with approximately half of the accepted tokens originating from Child-$k$ nodes (where $k>1$). These tokens are typically not taken into account in SSD. Comparing TSD and GSD, we can see that GSD slightly increases the acceptance rate for tokens positioned as Child-1.
Figure~\ref{fig:ap2} shows how varying the $\theta_{sib}$ threshold impacts the acceptance rate for tokens at each position. A higher $\theta_{sib}$ corresponds to more stringent pruning, resulting in fewer sibling nodes being retained. We can see a clear negative correlation between the increase of $\theta_{sib}$ and the acceptance rate for tokens at latter positions.

\end{document}